\documentclass[10pt]{article}
\usepackage[letterpaper]{geometry}
\usepackage{hicss}
\usepackage{times}
\usepackage[none]{hyphenat}
\usepackage{url}
\usepackage{latexsym}
\usepackage{indentfirst}
\usepackage{graphicx}
\graphicspath{{images/}}
\usepackage[
    style=apa,doi=false,isbn=false
  ]{biblatex}
\newcommand{\uproman}[1]{\uppercase\expandafter{\romannumeral#1}}

\usepackage{enumitem}
\newlist{questions}{enumerate}{2}
\setlist[questions,1]{label=\textbf{RQ\arabic*:},ref=RQ\arabic*}
\setlist[questions,2]{label=(\alph*),ref=\thequestionsi(\alph*)}

\newlist{hypotheses}{enumerate}{2}
\setlist[hypotheses,1]{label=\textbf{H\arabic*:},ref=H\arabic*}

\usepackage{multirow} 
\usepackage{rotating} 
\usepackage{rotfloat}

\usepackage{diagbox}

\usepackage[hidelinks]{hyperref}

\usepackage{tabularx}  
\usepackage{cleveref}
\usepackage{url}

\newcolumntype{b}{X}
\newcolumntype{s}{>{\hsize=.1\hsize}X}

\title{Deep Domain Adaptation for Detecting Bomb Craters in Aerial Images}

\author{Marco Geiger \\
  Karlsruhe Institute\\ of Technology (KIT) \\
  {\underline{marco.geiger@alumni.kit.edu}} \\\And
  Dominik Martin \\
  Karlsruhe Institute\\ of Technology (KIT) \\
  {\underline{dominik.martin@kit.edu} }\\ \\\And
  Niklas Kühl \\
  Karlsruhe Institute\\ of Technology (KIT) \\
  {\underline{niklas.kuehl@kit.edu} }\\}

\date{}

\begin{document}
\maketitle
\begin{abstract}
\label{sec:abstract}
The aftermath of air raids can still be seen for decades after the devastating events. Unexploded ordnance (UXO) is an immense danger to human life and the environment. Through the assessment of wartime images, experts can infer the occurrence of a dud. The current manual analysis process is expensive and time-consuming, thus automated detection of bomb craters by using deep learning is a promising way to improve the UXO disposal process. However, these methods require a large amount of manually labeled training data. This work leverages domain adaptation with moon surface images to address the problem of automated bomb crater detection with deep learning under the constraint of limited training data. This paper contributes to both academia and practice (1) by providing a solution approach for automated bomb crater detection with limited training data and (2) by demonstrating the usability and associated challenges of using synthetic images for domain adaptation.
\end{abstract}

\section{Introduction}
\label{subsec:Introduction}
For Germany in particular, unexploded bombs from World War \uproman{2} air raids are a problem that is still present. It is estimated that 10-20 \% of allied aerial bombs have failed to detonate \autocite{Harrabi2019}.
Today, tens of thousands of duds lie in the ground in Germany, posing a significant threat to the environment, buildings, animals and ultimately human lives \autocite{Dittrich2021}. However, unexploded ordnance (UXO) is not systematically searched for but only on a case-by-case basis. Each time a building application is submitted, a visual inspection of historical war images is carried out. By looking at aerial images taken from surveillance planes after an air raid, skilled experts assess whether a bomb is potentially unexploded in the area at hand \autocite{Dittrich2021}.
Since the analysis of the historical imagery can only be done by skilled staff and is mentally demanding, i.e., it cannot be carried out for multiple consecutive hours, the entire process is costly \autocite{Mayer2021}.

Prior research suggests using deep learning (DL), specifically convolutional neural networks (CNNs) to detect bomb craters on historical aerial imagery to partially automate or assist the task \autocite{Clermont}. CNNs trained on a large number of labeled images produce visual representations that may be utilized for the automatic detection of bomb craters. However, obtaining labeled data is a time-consuming and expensive procedure \autocite{Murez2018}. Qualified experts must identify and label the craters to create suitable training samples. Therefore, this paper challenges an approach to detect bomb craters in aerial wartime images with mitigating the limited training data issue by domain adaptation (DA) techniques creating synthetic training data through input data from other domains.

Three research questions are derived from the problem statement and motivation:

\begin{questions}[topsep=0pt,itemsep=-1ex,partopsep=1ex,parsep=1ex]
    \item To what extent can DL support the UXO disposal process?
    \item How can the problem of missing training data in bomb crater detection be solved?
    \item Can DA be used to improve the detection of craters in the UXO disposal process?
\end{questions}

Generally, the present work aims to evaluate to what extent DL can support the UXO disposal process and whether the problem of limited training data can be mitigated by applying DA techniques.

\section{Research Design} 
\label{subsec:RD}
We follow the Design Science Research (DSR) paradigm \autocite{Gregor2007, Peffers2007}, which aims to guide the creation of innovative artifacts solving practical problems with a sound methodological framework. Existing theory can serve as justificatory knowledge supporting the design and rigorous construction and evaluation of the artifact \autocite{Gregor2007}. To ensure this, we base our research on the three-cycle view proposed by \textcite{Hevner2007}. Thus, the goal of this article is to create an artifact (design cycle) that solves the introduced problem of bomb crater detection with limited training data (relevance cycle) while drawing on justificatory knowledge from theory (rigor cycle).

To structure the application of DSR, we guide our design through the six-step process of \textcite{peffers_process}. The process starts with the \emph{identification of the problem}, which is outlined in \Cref{subsec:Introduction} and is part of the relevance cycle. 
Subsequently, the \emph{objective definition} is based on relevant scientific foundations from literature (cf. \Cref{sec:TheoraticalFoundations}) and thus is part of the rigor cycle. We start the \emph{design \& development} (design cycle) step with a conceptual design suggestion leveraging synthetic images of bomb craters based on moon craters. Next, the design suggestion is instantiated by means of an artifact in the form of a model \autocite{Peffers2012}.

The artifact's evaluation is described in two evaluation episodes according to \textcite{Venable2016}. In a first artificial episode, a technical experiment is carried out (cf. \Cref{sec:TechnicalExperiments}), using four different datasets for training the crater detection model. In a second episode (cf. \Cref{sec:Discussion}), a more naturalistic evaluation is conducted considering image quality issues and a combination of different models for better inspection quality in practice.

On the one hand, this research contributes to the application domain by evaluating the potential of DL and DA to improve crater detection in the UXO disposal process under the constraint of limited training data availability. In particular, the use of synthetic data based on similar geological artifacts (i.e., lunar craters) using DA is a novelty.
On the other hand, the present work provides insights into the challenges and caveats of DA for synthetic image creation, thus, contributing to DL and DA research. Today, the practical use of DA for real-world problems is still insufficient due to the novelty of the research area---but promises enormous potential. Thus, this article represents a specific instance that aims to contribute to the applicability of DA in general.


\section{Theoretical Foundations}
\label{sec:TheoraticalFoundations}
A \emph{Generative Adversarial Network (GAN)} is a learning algorithm that was initially proposed by \textcite{Goodfellow} and is characterized by training a pair of neural networks (generator and discriminator) in an adversarial fashion. The goal of the generator is to learn the distribution of real data, whereas the discriminator's goal is to accurately identify whether the input data is drawn from real or artificial data produced by the generator. Generator $G$ generates samples $G(z)$, where $z$ is a random variable. Discriminator $D$ receives real samples $x$ as well as transformed data $G(z)$ and maps the given received data to a probability that the input is from real data $x$ $(P=1)$ or from $G(z)$ $(P=0)$. Thus the purpose of $D$ is to make correct classifications on the given input ($D(x,G(z)) \rightarrow (0,1)$), while $G$ tries make the performance on the generated data ($G(z)$) consistent with the performance on real data $D(x)$ \autocite{Creswell2018}. The discriminator will be maximally confused if the generator distribution $p_g(x)$ can completely match the real data distribution $p_{data}$.

\emph{Image-to-Image Translation (I2I)} is the task of transferring images from a particular source domain $X$ to a target domain $Y$ while preserving the content representation \autocite{Liu, Pang2021}. In the unsupervised case, two independent sets of images exist and no paired examples. Consider the input images $x$ from source domain $X$ and the images $y$ from target domain $Y$. To carry out the I2I translation, a mapping $G_{X \rightarrow Y}$ is needed to generate image $x \in X$ that cannot be distinguished from a target domain image $y \in Y$ given a source image $x \in X$ \autocite{Pang2021}.


\emph{CycleGAN} is an architecture that extends the idea of using generative models for unsupervised I2I. The mapping $G: X \rightarrow Y$ for the I2I translation does not imply that an individual input $x$ and output $\hat{y}$ can be associated \autocite{Goodfellow, Murez2018, Zhu2017}. For an image, this would mean that the I2I translation does not preserve the visual structure of the original image \autocite{Creswell2018}. Therefore \emph{cycle} consistency is introduced to the translation process. This means that an input image $x$ is transformed to the target domain and generates an image $\hat{y}$, which generates the original image $x$ when transformed back into the source domain. To do this, CycleGAN uses two translators $G: X \rightarrow Y$ and $F: Y \rightarrow X$, which are inverse of each other and their mappings are bijections. To ensure that an individual input $x_i$ is mapped to a desired output $y_i$, the space of possible mapping functions is reduced by introducing a \emph{cycle consistency loss} \autocite{Sundaram2010}. The loss encourages $F(G(x)) \approx x$ and $ G(F(y)) \approx y$.

\emph{You only look once (YOLO)} is a supervised DL algorithm for object detection. YOLO is a one-stage object detection approach that treats the detection problem as a continuous regression from the input image to the final detections. YOLO uses a CNN to extract features from the input image. The resulting feature maps are used in the final layers to determine the positions (bounding boxes) and classes of objects. Since YOLO is a supervised approach, the training requires labeled data \autocite{Redmon}.

\section{Related Work}
\label{sec:RelatedWork}
Related research work can be divided into two types.
On the one hand, related research pursues the same goal of crater detection, 
and on the other hand, related work uses the same DA approach.

Concerning the goal of crater detection, the most common applications and algorithms exist in geoscience and are concerned with exploring the geology of the moon and mars. Many image data approaches use circular pattern recognition \autocite{Salamuniccar2011}, while others try to exploit shadow regions of craters \autocite{Urbach2009}. In addition, \textcite{Jung2005} approach the detection of craters with a multi-stage template matching approach. Other research tries to detect craters directly from Digital Elevation Model (DEM) information \autocite{Wang2019}. The criticism of the presented methods is that high accuracy of crater detection can only be achieved on specific data, certain planetary surfaces and for certain craters \autocite{Stepinski2012}.

Recently Machine Learning approaches, especially CNNs, have shown outstanding results on the task at hand. The research work based on CNN use the networks for either classification of craters \autocite{Cohen2016}, crater detection \autocite{Benedix, Emamib} or crater segmentation \autocite{Lee2019, Silburt2019, DeLatte}.
\textcite{Finkelstein} and \textcite{Silburt2019} show that a transfer of a trained model on lunar craters also yields good results when applied to martian craters. 

There is far less research on detecting bomb craters in aerial war images. \textcite{Merler2005} and \textcite{Brenner2018} use a sliding window approach to create search windows for crater candidates, while the latter use a  CNN to classify crater candidates and the first  use a variant of AdaBoost. In contrast to the research presented above, \textcite{Kruse2018} do not aim to provide a complete detection of bomb craters but rather impact maps, indicating the contamination of a specific area. \textcite{Clermont} use a blob detector to select crater candidates classified by a CNN. The results show an incompleteness in automated crater detection for several datasets.

Regarding Domain Adaptation, a distinction can be made between shallow and deep DA methods \autocite{Wang}. The most prevalent shallow DA algorithms can be divided into two groups: instance-based DA \autocite{Bruzzone, Chu} and feature-based DA \autocite{Gheisari2015a, Long2013}. 
The first class reduces the disparity by reweighing the source samples, whereas the second class learns a common shared space in which the distributions of the two datasets are matched \autocite{Wang}. Recently, neural networks for DA have shown significant improvements. 

These so-called deep DA methods use neural networks and can be categorized into methods using feature adoption and methods using generative models \autocite{You2019}. Feature adoption techniques use several approaches, for example, minimizing the maximum mean discrepancy of deep features across domains \autocite{Long2015}. The research works of \textcite{Ganin2017, Bousmalis2016, Hoffman2017} all use an adversarial learning paradigm to train a domain classifier that differentiates features  from source and target domains and pushes the feature extractor to confuse that domain classifier. Besides the feature adoption methods presented, generative methods based on GANs show impressive results by creating synthetic training samples \autocite{Murez2018, Huang2018, Liu2017a, Sankaranarayanan2017, Bousmalis2016}. These generative methods build on I2I translation networks such as CycleGAN \autocite{Zhu2017}, AugGAN \autocite{Huang2018} and DiscoGAN \autocite{Kim2017}.

\begin{figure*}[h]
	\centering
	\includegraphics[width=.85\textwidth]{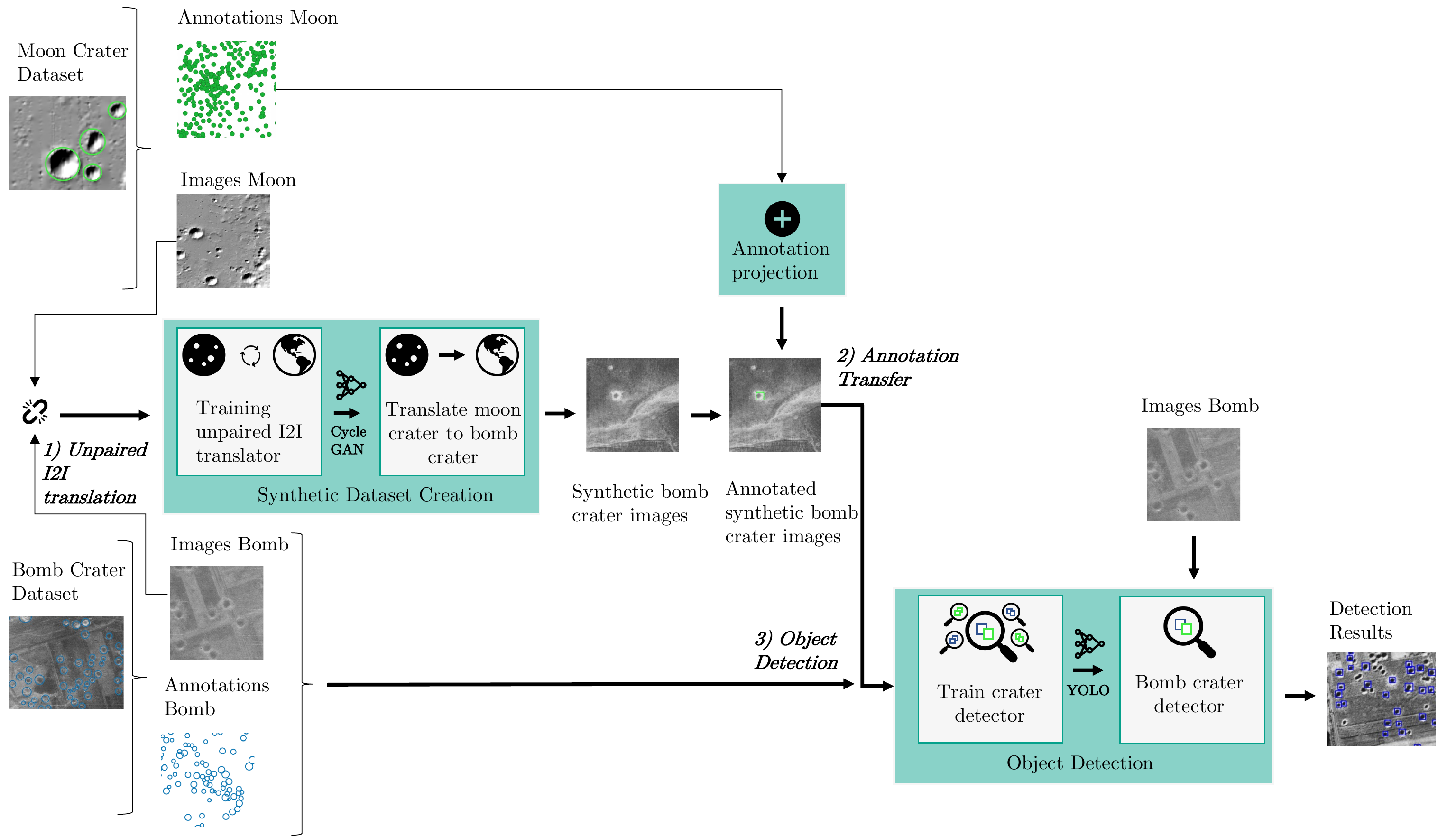}
	\caption{Concept of the design artifact}
	\label{fig:design_overview}
\end{figure*}

The CycleGAN approach, in particular, serves as a basis for the further development of DA techniques, such as CyCada \autocite{Hoffman2018} or SBADA-GAN \autocite{Russo2018}. The research of Schutera et al. \autocite{schutera}, Lin \autocite{Lin2019} and Arruda et al.\autocite{Arruda2019} shows improved object detection results for autonomous driving based on generative models that translate daylight to night images.

\section{Design Cycle}
\label{sec:DesignCycle}

As a first design step, we suggest a tentative design in the form of a conceptual framework based on identified requirements (cf. relevance cycle) and foundational methods (cf. rigor cycle). \Cref{fig:design_overview} depicts the conceptualized artifact steps. The presented artifact creates synthetic images of bomb craters based on moon crater images. It adapts the source domain moon crater to the target domain bomb crater in order to create training data, respectively improving a bomb crater detection model.

The concept envisages training an \emph{unpaired I2I translation} model that is able to transform images of the source domain moon crater to the visual appearance of the target domain bomb crater. The output is a synthetic image dataset whose images have the same contours and shapes as the source lunar crater images.
Therefore, in the \emph{annotation transfer} step, the labels are projected onto the artificially generated images. Subsequently, in the last step \emph{object detection}, the actual bomb crater dataset and the synthetic dataset are used to evaluate the artifact.
This is carried out by training several crater detection models and comparing them based on their detection quality on unseen test images.

\subsection{Dataset Analysis and Preparation}
\label{sec:DatasetAnalysisandPreparation}
\begin{figure}[h]
	\centering
	\includegraphics[angle=270, width=0.8\linewidth]{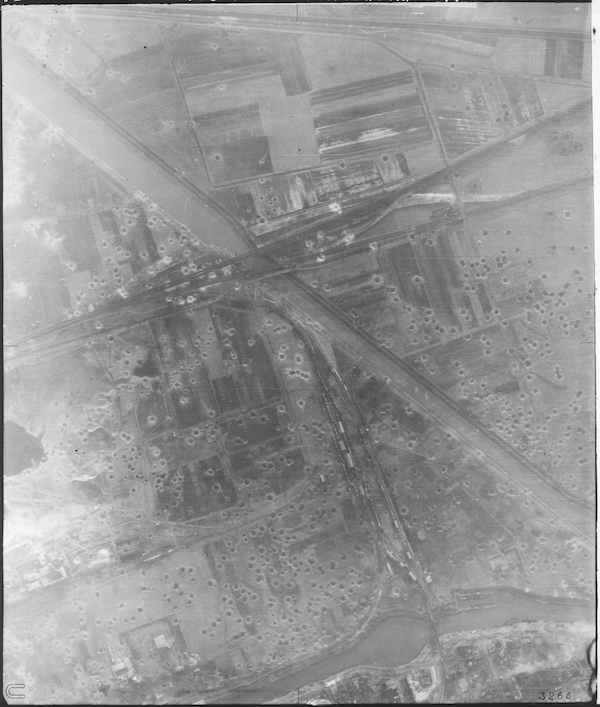}
	\caption{Exemplary bomb crater image $t_i$}
    \label{fig:bomb_crater}
\end{figure}

\begin{figure}[h]
	\centering
	\includegraphics[width=0.8\linewidth]{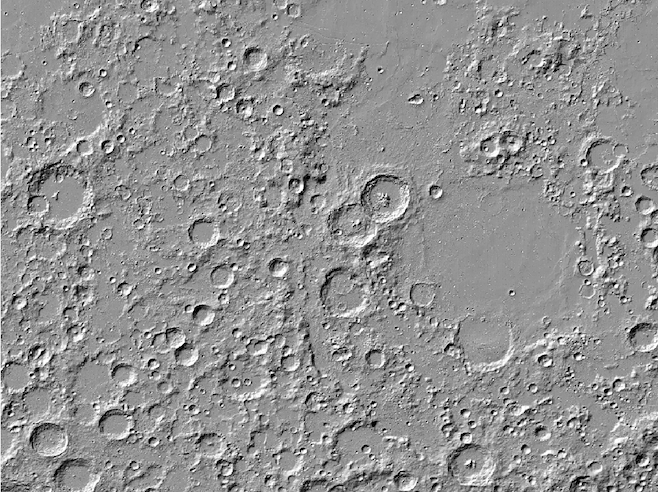}
	\caption{Exemplary moon crater image $s_i$}
	\label{fig:moon_crater}
\end{figure}

To instantiate and evaluate the suggested artifact, we use real-world bomb crater images (cf. \Cref{fig:bomb_crater}; dataset $T$ representing the target domain), and images of the lunar surface (cf. \Cref{fig:moon_crater}; dataset $S$ representing the source domain). The aerial images $t \in T$ were taken during surveillance flights after allied air raids during WW\uproman{2}. Domain experts from Lower Saxony's explosive ordnance disposal service labeled the images manually to create the annotation set $C$. 

Dataset $S$ is created from a single, geo-referenced image of the moon surface \autocite{NASA2021}. As annotations for $S$, we introduce set $M$, derived from the moon crater database v1 Robbins \autocite{NASA2021a}, which contains around 1.3 million lunar crater positions \autocite{Robbins2019}.

Several preparation steps are necessary to use the datasets mentioned above as input for the I2I translation and the training of an object detection model. These steps are shown in \Cref{tab:data_preparation}.
For the image data, this includes a Region of Interest (ROI) cropping, mitigation of noise (e.g., clouds, grey and black shadows) through contrast limited adaptive histogram equalization \autocite{Pizer1987} and a tiling algorithm \autocite{Unel2019} to generate suitable input sizes for feeding the data into a neural network.
For the label data, the processing includes transforming the global coordinates onto the generated image tiles and selecting the lunar crater labels based on diameter and visual appearance using a pixel-based heuristic.
The final results are the processed datasets $S'$, $T'$, $M'$ and $C'$, which are used to create synthetic images of the target domain bomb crater.

\begin{table}[h]
	\centering\small
	\begin{tabularx}{8cm}{s b s}
		\hline
		Input & Processing Steps & Output\\
		\hline
		$S$ & Tiling & $S'$ \\
        $T$ & ROI Crop; Quality Improvement; Tiling & $T'$\\
        \hline
        $M$ & Transform coordinates; Filter craters based on diameter; Prune crater set pixel-based; Project labels & $M'$\\
        $C$ & Transform coordinates; Project labels & $C'$ \\
		\hline
	\end{tabularx}
	\caption{Data Preparation}
	\label{tab:data_preparation}
\end{table}

\subsection{Synthetic Images Creation}
\label{sec:SyntheticImagesCreation}
We use the cycleGAN \autocite{Zhu2017}, to carry out an I2I translation. The goal is to create a cycle consistent mapping from the moon crater source domain to the target bomb crater domain $G: S' \rightarrow T'$ such that the synthetic output image $\hat{t} = G (s')$, $s' \in S'$ cannot be distinguished from the original bomb crater image $t' \in T'$.

According to the DSR framework, we develop the artefact in an iterative approach. We analyze our findings during the development process and try to adapt the process to improve our results. As there are no meaningful quantitative metrics for assessing a trained cycleGAN model \autocite{Zhu2017}, we visually analyze the produced synthetic images. We achieve significantly better results by adapting hyperparameters (e.g., batch size, epochs, image size), but especially by filtering the input data based on image similarity and present distortions. Moreover, it became apparent that the craters of the transformed images $\hat{t'} \in \hat{T'}$ look better if the target domain images constantly show craters, not only landscape.

\begin{figure}[h]
	\centering
	\includegraphics[width=0.48\textwidth]{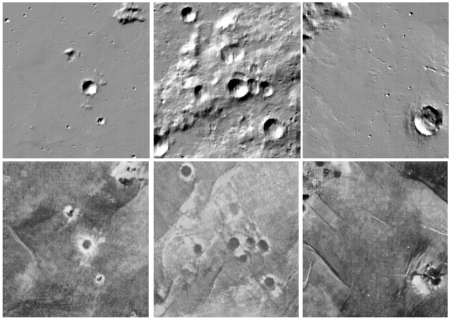}
	\caption{I2I translations from moon (upper row) to bomb crater domain (lower row)}
	\label{fig:fake_images}
\end{figure}

\begin{figure}[h]
	\centering
	\includegraphics[width=0.48\textwidth]{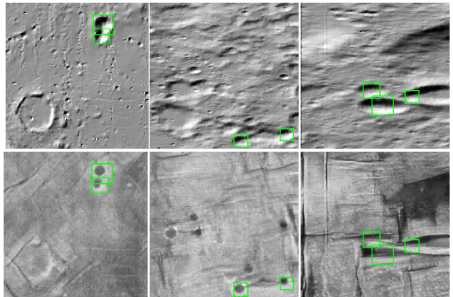}
	\caption{Annotations of $M'$ projected on tiles of $\hat{T'}$. First column shows good examples; second column craters without annotations; third column the opposite case.}
	\label{fig:examples_annotations}
\end{figure}

After generating the artificial dataset $\hat{T'}$, the annotations $M'$ are projected onto the synthetic images. Unfortunately, despite the preprocessing of set $M$, the annotations still have inaccuracies. This is primarily due to the fact that previously visible craters of the original images in $S'$ are no longer recognizable as craters after the I2I translation.
\Cref{fig:examples_annotations} shows good and bad examples from set $\hat{T'}$ with corresponding annotations from $M'$.

\subsection{Technical Experiments}
\label{sec:TechnicalExperiments}
To evaluate the artifact, a technical experiment \autocite{Peffers2012} is carried out, which assesses the quality improvement of automated crater detection by leveraging synthetic image data.
The object detection network ScaledYolov4 \autocite{Wang2021} is trained on the bomb crater data dataset $T'$ with annotations $C'$ to serve as a reference. To compare this base case scenario against our proposed approach, we pose three hypotheses to show the effect of DA:

\begin{hypotheses}[topsep=0pt,itemsep=-1ex,partopsep=1ex,parsep=1ex]
    \item A model trained on moon crater images leads to equally good detection results as a model trained on real bomb crater images.
    \item A model trained on synthetic bomb crater images leads to equally good detection results as a model trained on real bomb crater images.
    \item Adding synthetic bomb crater images to the original bomb crater training data leads to improved detection results.
\end{hypotheses}

The dataset $T'$ is split into three subsets, a training, a validation and a test set.
\Cref{tab:YOLO_model_trainings} shows an overview of the models based on hypotheses H1-3.

\begin{table}[h]
    \centering\small
    \begin{tabularx}{\linewidth}{XXXX}
    \hline
    Model & Training set & Validation set & Test set \\
    \hline
    $bomb$ & $T'_{train}$ & \multirow{4}{*}{$T'_{val}$} & \multirow{4}{*}{$T'_{test}$} \\
    $moon$ & $S'$ & & \\
    $synthetic$ & $\hat{T'}$ & & \\
    $combined$ & $T'_{train}\:\&\:\hat{T'}$ & & \\
    \hline
    \end{tabularx}
    \caption{Overview of models and used datasets}
    \label{tab:YOLO_model_trainings}
\end{table}

We use common metrics for evaluating the models, particularly the \emph{precision}, \emph{recall} and the $mAP_{0.5}$ \autocite{padilla} metrics. The $mAP_{0.5}$ uses the average precision calculated for the 0.5 IoU threshold. For the crater detection task at hand, $mAP_{0.5}$ is considered sufficient because a detection with $IoU > 0.5$ is accurate enough to indicate the presence of a crater, which human experts can use in the UXO disposal process to classify duds \autocite{Kruse2018}.

\subsection{Evaluation}
\label{sec:Evaluation}
\Cref{tab:evaluation_YOLO} illustrates the introduced metrics for the different models depicted in \Cref{tab:YOLO_model_trainings}.
Looking at the different train losses as well as the $mAP_{train}$ scores, it can be seen that all models achieve solid performance on their respective training data. However, the validation losses and the $mAP_{val}$ scores show a significantly worse performance on the validation set $T'_{val}$.
Looking at the metrics $mAP_{val}$ and $mAP_{test}$, it can be seen that there is a poor performance on the bomb crater data. The model detects no craters in aerial wartime images, thus, hypothesis \emph{H1} can be falsified.
Moreover, hypothesis \emph{H2} can be rejected since the synthetically generated dataset $\hat{T'}$ leads to a worse $mAP_{test}$ compared to the $model_{bomb}$.
Hypothesis \emph{H3} cannot be verified or falsified with certainty based on the quantitative metrics. The $mAP_{test}$ is higher for the $model_{combined}$ than for the $model_{bomb}$, but it is rather a slight improvement.

\begin{table}[h]
	\centering\small
	\begin{tabular}{l ccccc}
		\hline
		Metric & $bomb$ & $moon$ & $synthetic$ & $combined$\\
		\hline
		$train\:loss$ & 1.645 & 3.098 & 2.312 & 1.411 \\
        $validation\:loss$ & 5.589 & 6.767 & 5.698 & 5.136 \\
        $mAP_{train}$ & 0.717 & 0.93 & 0.837 & 0.876 \\
        $mAP_{val}$ & 0.245 & 0 & 0.046 & 0.239  \\
        $mAP_{test}$ & 0.099 & 0 & 0.043 & 0.103  \\
        $precision_{test}$ & 0.188 & 0 & 0.083 & 0.191 \\ 
        $recall_{test}$ & 0.288 & 0 & 0.164 & 0.278  \\
		\hline
	\end{tabular}
	\caption{Performance metrics}
	\label{tab:evaluation_YOLO}
\end{table}

\section{Discussion}
\label{sec:Discussion}
Overall, the quantitative metrics in \Cref{tab:evaluation_YOLO} indicate a poor detection quality of all models on the available bomb crater images. There seems to be strong overfitting for each model, which could be due to the high complexity, i.e., the high number of layers of ScaledYOLOv4 \autocite{Bejani2019} or the fact that the amount of data generally seems sufficient, but the images often show the same geographical areas, thus, the same impact craters. However, these quantitative metrics cannot be considered a single source of truth. 

\begin{figure}[t]
	\centering
	\includegraphics[height=275pt, width=0.48\textwidth]{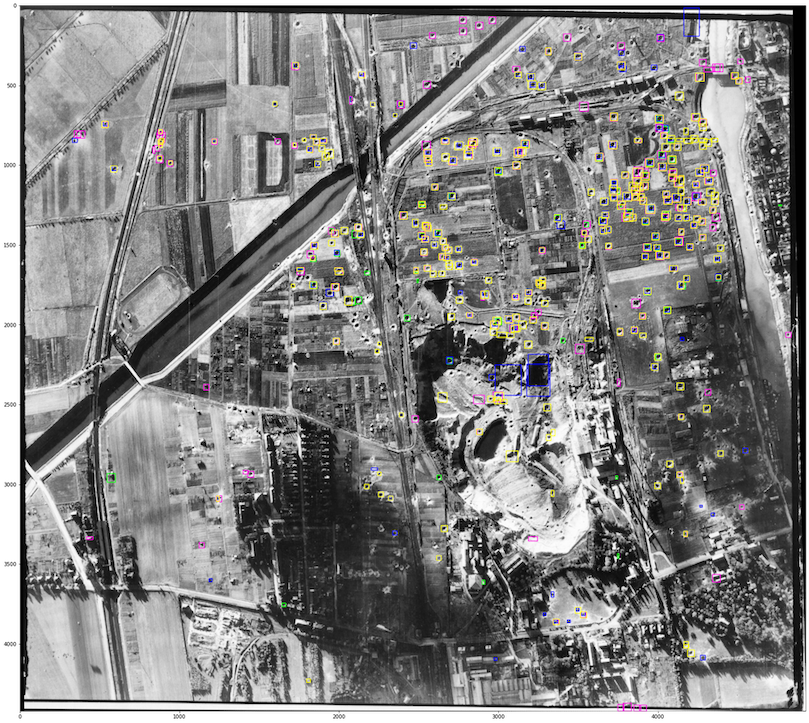}
	\caption{Detections of different models as colorized bounding boxes}
	\label{fig:test_ensemble}
\end{figure}

Visual inspection of the test dataset $T'_{test}$ reveals that some recognizable bomb craters are not labeled. Moreover, the test set, in particular contains images of poor quality, which stems from geographical clusters of the bomb crater imagery, limiting the training, validation, and test data selection. These constraints are limitations of the present work and impair the reliability of the quantitative evaluation. Therefore, a visual analysis of the detection results is carried out to generate conclusive findings. \Cref{fig:test_ensemble} shows an image that is not included in any of the introduced datasets and can serve as a basis for qualitative analysis. Crater detections for each trained model are drawn on the image and indicated by colored bounding boxes. Green stands for $model_{moon}$, blue for $model_{bomb}$, pink for $model_{synthetic}$ and yellow for $model_{combined}$.

The $model_{moon}$ does not show good detection results both qualitatively and quantitatively. This means that a direct transfer of a crater detection model from the \emph{source domain} moon craters to the \emph{target domain} bomb craters is not possible and there is a need for applying DA.

Concerning the synthetic data, visual observation reveals a high number of False-Positive (FP) detections from the $model_{synthetic}$, indicated by the significantly lower $precision_{test}$ value compared to the $model_{bomb}$. The $model_{synthetic}$ often detects trees, houses or other objects as craters since these represent a collection of darker pixels surrounded by brighter pixels. This likely happens because the synthetic images used for training mainly show craters and no other landscape objects. It seems that the model does not explicitly learn the difference between these objects in the training process. Generally, it can be observed that the performance of the $model_{synthetic}$ is significantly worse compared to the $model_{bomb}$ even in the $recall_{test}$ metric, which means that no equivalent detection quality is achieved even if the FP detections are ignored. However, FP detections are not as problematic in the overall context of the UXO disposal process, since the human experts could quickly identify erroneously detected craters visually.

The barely improved performance of the $model_{combined}$ compared to the $model_{bomb}$ seems unusual at first since the visual similarity of the synthetic images to the original bomb crater images is given, as shown in \Cref{fig:fake_images}. Due to the more extensive training set and higher variance of craters, improved results were expected. Reasons for the lack of this improvement could be an insufficient annotation quality in $M'$ or still an inadequate quality of the fake images in $\hat{T'}$ as the quality could only be assessed visually. 

Although synthetically generated training data does not significantly affect the quality of object detection, it is evident that the models perform differently in different image regions. Looking at \Cref{fig:test_ensemble}, the upper edge of the image is particularly striking, as the $model_{synthetic}$ (indicated in pink) detects craters that the other models do not recognize. In addition, the sandpit in the middle of the image is conspicuous, as only the $model_{synthetic}$ and the $model_{combined}$ produce suitable detections. 
This suggests that models trained with synthetic images are superior in detecting craters with a brighter background. Moreover, the partially low metrics originate from the existence of crater fields on aerial wartime images. It frequently happens that the trained models do not detect all craters of a crater field. However, this is not problematic in practical use since humans can quickly recognize the entire crater field based on the detected craters of the automated detection. 
The detection of each individual crater is not necessarily decisive to improve the process \autocite{Kruse2018}.

Furthermore, other advantages arise despite the low detection quality. 
Detecting outlier craters within the UXO disposal process is especially difficult for human experts.
Especially for these crater types, automatic crater detection may offer added value, although the overall detection performance is relatively low. While the quality of crater detection, even with the addition of synthetically generated data, seems insufficient for the problem of automatic crater detection presented in \Cref{subsec:Introduction}, the combination of detection results from the different models might be a promising approach to provide overall good detection results. Especially because images $t'_i \in T'_{test}$ partially overlap and show the same geographical area, so that multiple models can use different images for detecting a specific crater. This is a meaningful finding of the conducted experiment and suggests the application of ensemble techniques \autocite{Murphy} to generate a higher overall crater detection quality. Therefore, even a moderate detection quality may have an economic benefit in terms of accelerating the process.

\section{Conclusion and Outlook}
\label{sec:conclusion}
This research used the DSR methodology to address the problem of automated bomb crater detection with limited training data in the UXO disposal process.
For this purpose, a DA approach was conceptualized to use publicly available moon craters for detecting bomb craters. More specifically, an artifact based on state-of-the-art GAN research was created to generate synthetic training data for an object detection network. This artifact was evaluated by an experiment investigating the detection quality of several bomb crater detection models trained on different datasets.

Regarding \emph{RQ1}, the experiment showed that a fully automated detection of bomb craters is not possible due to insufficient detection quality. This finding is in line with the recommendations of previous research \autocite{Clermont}. However, even moderate detection quality can support the manual UXO detection task. Concerning \emph{RQ2}, it can be stated that missing training data of the bomb crater domain can be generated synthetically with visually good quality. However, this involves considerable effort. Furthermore, it is shown that synthetically generated bomb crater data cannot equally replace original images. Finally, with regard to \emph{RQ3}, it can be said that the use of DA does not directly improve a single bomb crater detection model. However, combining the detections from several models achieves better results.

This work highlights challenges and requirements when using DA based on a practical use case. It should be emphasized that I2I translation in combination with object detection has difficulties regarding the consistent use of labels, because small objects may disappear. This issue is accompanied by increased efforts in preparing datasets, as source datasets usually cannot be used for DA without further adjustments.

Since the application of DA through synthetic data generation is still at an early stage of research, more practical cases should be contributed to generalize the generated findings. Moreover, it is crucial to develop criteria to assess when datasets or practical use cases are suitable for DA. In general, indicators need to be developed for assessing which type of DA is suitable for a particular task or domain.



\printbibliography

\end{document}